\documentclass{article}

\usepackage[utf8]{inputenc}
\usepackage{fontenc}
\usepackage{amsmath}
\usepackage{amsfonts}
\usepackage{amssymb}
\usepackage{graphicx}
\usepackage{booktabs}
\usepackage{multicol}
\usepackage[margin=1in]{geometry}
\usepackage{hyperref}
\usepackage{url}
\usepackage{authblk}
\usepackage{nicefrac} 
\usepackage{microtype}
\usepackage{doi}
\usepackage{caption}
\usepackage{subcaption}

\hypersetup{
    colorlinks=true,
    linkcolor=blue,
    filecolor=magenta,      
    urlcolor=cyan,
    pdftitle={Reactive Transformer},
    pdfpagemode=FullScreen,
}

\title{
    \textbf{Reactive Transformer (RxT) - Stateful Real-Time Processing for Event-Driven Reactive Language Models}
}
\author{Adam Filipek (adamfilipek@rxai.dev)}
\affil{Reactive AI (https://rxai.dev)}
\date{October 2025}

\begin{document}
\maketitle

\begin{abstract}
The Transformer architecture has become the de facto standard for Large Language Models (LLMs), demonstrating remarkable capabilities in language understanding and generation. However, its application in conversational AI is fundamentally constrained by its stateless nature and the quadratic computational complexity ($O(L^2)$) with respect to sequence length $L$. Current models emulate memory by reprocessing an ever-expanding conversation history with each turn, leading to prohibitive costs and latency in long dialogues. This paper introduces the Reactive Transformer (RxT), a novel architecture designed to overcome these limitations by shifting from a data-driven to an event-driven paradigm. RxT processes each conversational turn as a discrete event in real-time, maintaining context in an integrated, fixed-size Short-Term Memory (STM) system. The architecture features a distinct operational cycle where a generator-decoder produces a response based on the current query and the previous memory state, after which a memory-encoder and a dedicated Memory Attention network asynchronously update the STM with a representation of the complete interaction. This design fundamentally alters the scaling dynamics, reducing the total user-facing cost of a conversation from quadratic ($O(N^2 \cdot T)$) to linear ($O(N \cdot T)$) with respect to the number of interactions $N$. By decoupling response generation from memory updates, RxT achieves low latency, enabling truly real-time, stateful, and economically viable long-form conversations. We validated our architecture with a series of proof-of-concept experiments on synthetic data, demonstrating superior performance and constant-time inference latency compared to a baseline stateless model of comparable size.
\end{abstract}

\section{Introduction}

The introduction of the Transformer architecture \cite{vaswani2017attention} marked a watershed moment in Natural Language Processing (NLP). Its self-attention mechanism proved exceptionally effective at capturing long-range dependencies, leading to the rise of Large Language Models (LLMs) like GPT and BERT, which now dominate the field. These models excel at a wide range of tasks by processing vast sequences of text in a single forward pass.

However, this operational paradigm reveals a critical flaw when applied to conversational tasks. Transformers are inherently stateless; each input sequence is processed in isolation. To maintain conversational context, models are forced to re-ingest and re-process the entire history of the dialogue with every new user message. This "brute-force" approach to context management results in a computational cost that scales quadratically with the number of conversational turns, rendering long-running dialogues economically impractical and leading to significant latency issues. While ever-larger context windows (up to millions of tokens) are being explored, they do not solve the underlying scaling problem but merely postpone it at an escalating cost.

Moreover, that stateless history reprocessing is not only extremely inefficient, but also fundamentally wrong for the awareness and real AGI, and is not natural - that's not how humans think. We don't need to remind all day or week history in order to remember what we were doing 10 minutes ago. Awareness also require continuous processing, but with stateless LLMs it's more expensive and slower on each step, and is limited to model's context, so they are not compatible. Our "Reactivity Hypothesis" states that "real awareness and AGI models require continuous, stateful, real-time processing" - LLMs are not fulfilling any from the requirements, while Memory-Augmented Neural Networks fulfill them only partially.

This paper argues that the path forward for conversational AI requires a fundamental paradigm shift from data-driven stateless processing to event-driven, stateful computation. We introduce the \textbf{Reactive Transformer (RxT)}, an architecture designed from the ground up for stateful real-time dialogue. RxT treats each user query as an event, processes it against an internal memory state, generates a response, and then asynchronously updates its memory. This approach offers two primary contributions:

\begin{enumerate}
    \item \textbf{A Novel Event-Driven Architecture:} RxT redefines the roles of the encoder and decoder within a cyclical, asynchronous workflow. This decouples the user-facing task of response generation from the internal task of memory consolidation, drastically reducing perceived latency.
    \item \textbf{Linear Computational Scaling:} By maintaining a fixed-size internal memory and processing only the current interaction, RxT eliminates redundant computation. The total cost of a conversation scales linearly with the number of turns, making long, coherent dialogues computationally feasible.
\end{enumerate}

We posit that this architecture represents a crucial step away from simply scaling existing models and towards creating systems that can maintain a persistent, evolving understanding of an interaction, akin to a true conversational partner.

\section{Background - the evolution of conversation modeling}

The challenge of maintaining state and context in sequence modeling is not new. Our work builds upon a rich history of research, positioning RxT as a synthesis of prior concepts and a novel solution to their limitations.

\subsection{From Recurrence to Attention}
Early approaches to sequence modeling were dominated by Recurrent Neural Networks (RNNs), including Long Short-Term Memory (LSTM) and Gated Recurrent Unit (GRU) networks \cite{hochreiter1997long}. These models process sequences token-by-token, maintaining an internal hidden state that serves as a compressed memory of past information. While naturally stateful, RNNs suffer from the vanishing gradient problem and sequential processing bottlenecks, limiting their ability to capture very long-range dependencies and leverage parallel hardware \cite{bengio1994learning}. The Transformer architecture supplanted RNNs by replacing recurrence with a self-attention mechanism, allowing for parallel processing and superior modeling of long-distance relationships within a single sequence \cite{vaswani2017attention}. However, this came at the cost of abandoning the inherent inter-sequence statefulness of RNNs.

\subsection{Augmenting Neural Networks with Memory}
The concept of equipping neural networks with external, addressable memory was explored in early architectures like Neural Turing Machines (NTM) and Differentiable Neural Computers \cite{graves2014neural, graves2016hybrid}. These models coupled a neural network "controller" with a memory matrix, learning to perform read and write operations via differentiable attention mechanisms. While powerful, these systems were often difficult to train and did not see widespread adoption.

\subsection{System-Level Memory for LLMs}
With the dominance of stateless LLMs, memory has been reintroduced at the system level rather than at the architectural level. Retrieval-Augmented Generation (RAG) frameworks connect LLMs to external knowledge bases (e.g., vector databases) to provide factual grounding \cite{lewis2020retrieval}. Agentic frameworks such as LangChain orchestrate interactions between LLMs and various tools, including text-based memory modules. Although effective, these approaches treat the LLM as a black-box reasoning engine. The memory remains external, explicit, and subject to the same context window and reprocessing limitations of the underlying model - it is just a text, added to prompt before user's query/chat history, that could be only summarized, so its expressiveness and compression possibilities are weak. What's most important, LLM agents still operate in inefficient and incorrect (for awareness) data-driven full history processing paradigm.

\subsection{Alternative Architectures for Long Sequences}
To address the quadratic complexity of attention, alternative architectures have emerged. State Space Models (SSMs), such as S4 and its successor Mamba, exhibit linear or near-linear scaling with sequence length \cite{gu2021efficiently, gu2023mamba}. SSMs are inspired by control theory and use a recurrent mechanism that is highly parallelizable during training. However, their state is designed to capture dependencies \textit{within} a continuous sequence (intra-sequence), not to manage a persistent, addressable memory \textit{between} discrete conversational interactions (inter-sequence). In dialogue applications, they still rely on processing the full, growing history.

\subsection{Stateful Memory-Augmented Transformers}
More recently, efforts have been made to reintegrate memory directly into the Transformer architecture. The Stateful Memory-Augmented Transformer (MemBART) \cite{wu2022stateful} and it's predecessor Memformer \cite{wu2020memformer} are a notable examples. They augment a pre-trained encoder-decoder model with a memory module. However, its operational cycle is synchronous: the encoder processes the input query, reads from, and writes to memory, and only then passes the result to the decoder. This means that all memory operations contribute directly to the user-perceived latency, making it less suitable for real-time applications. Furthermore, they update their memory based solely on the input sequence, so previous answer still have to be included in input with query, making their processing not real-time (that require processing only current query/interaction), and initial processing phase is even slower. Stateful MATs were designed as memory extension for encoder-decoder transformers, extending their standard training methods with additional step. 

\subsection{Next generation of the stateful processing - Reactive Transformer}
Those stateful architectures were steps towards correct direction, with natural kind of processing and better results for conversational task than their stateless LLM counterparts, even with the limited expressiveness of their memory systems. However, they were released in 2020 and 2022, before the release of ChatGPT in 2023, which made the fundamentally wrong stateless "chat template" processing a default choice. Since then, mainstream research has been moving in this wrong direction, further extending the model contexts, leading to even greater inefficiencies and costs. The decision to choose and develop unnatural stateless LLMs, which have almost no advantages (only a smaller number of overall parameters) over the natural stateful processing,  has led to an unnecessary waste of enormous amounts of money and energy, especially for inference. It also pushed the research into directions like deep KV-cache optimizations, that in fact aren't even needed in stateful processing (however, Reactive Transformer is using KV-cache for self-attention and full pre-cache for memory cross-attention for even better performance). The argument, that access to all the history tokens is an advantage over fixed-size memory is weak - it's rather one of the biggest disadvantage of stateless modeling and the main source of model's hallucinations in long multi-turn dialogues. Even the biggest LLM models have very weak responses in long conversations, full of hallucinations that are mixing information from different time steps. On the opposite, fixed-size memory provides the most important information for the dialogue in each step, potentially limiting hallucinations.

The Reactive Transformer addresses the shortcomings of these prior works. Unlike LLMs, it is natively stateful. Unlike RNNs, it leverages the power of parallel attention within interactions. Unlike RAG, its memory is an integrated, implicit part of the model. And unlike synchronous MATs like MemBART, its asynchronous memory update is based on both query and answer, achieving unnoticeable latency, making it truly suitable for real-time processing - all the responses are generated instantly in almost the same time, no matter of number of previous messages. Reactive Transformer is not the extension for current architectures, but redefines processing paradigm and requires separately designed custom training stages, that are finally cheaper than LLM training, even when they are more advanced.

\section{The Reactive Transformer Architecture}

The Reactive Transformer is an encoder-decoder architecture designed around the \textbf{Event-Driven AI} paradigm. Instead of processing a monolithic data sequence (the entire conversation history), the model operates in a continuous loop, treating each query-response pair as a discrete \textbf{interaction}. Its core components are a Generator-Decoder, a Memory Encoder, and a Memory Attention network, which collectively manage a multi-layer Short-Term Memory (STM) state.

The operational flow of RxT fundamentally reverses the classic Transformer encoder-decoder pipeline and makes it cyclical:
\begin{enumerate}
    \item \textbf{Response Generation:} At interaction step $t$, the Generator-Decoder receives the user query $X_t$ and generates a response $Y_t$. This process is conditioned on the memory state from the previous interaction, $STM_{t-1}$, which is accessed via Memory Cross-Attention.
    \item \textbf{Asynchronous Memory Update:} After the response $Y_t$ has been generated and streamed to the user, the Memory Encoder processes a concatenation of the full interaction, $concat(X, Y)$, to produce a rich semantic representation, the Encoded Data ($ED_t$).
    \item \textbf{Memory Consolidation:} The Memory Attention network takes the previous memory state $STM_{t-1}$ and the Encoded Data $ED_t$ as input to compute the updated memory state, $STM_t$. This new state is then carried over to the next interaction, $t+1$.
\end{enumerate}

This asynchronous cycle ensures that the computationally intensive memory update process does not block the generation of the response, minimizing user-perceived latency.

\begin{figure}
    \centering
    \includegraphics[width=1\linewidth]{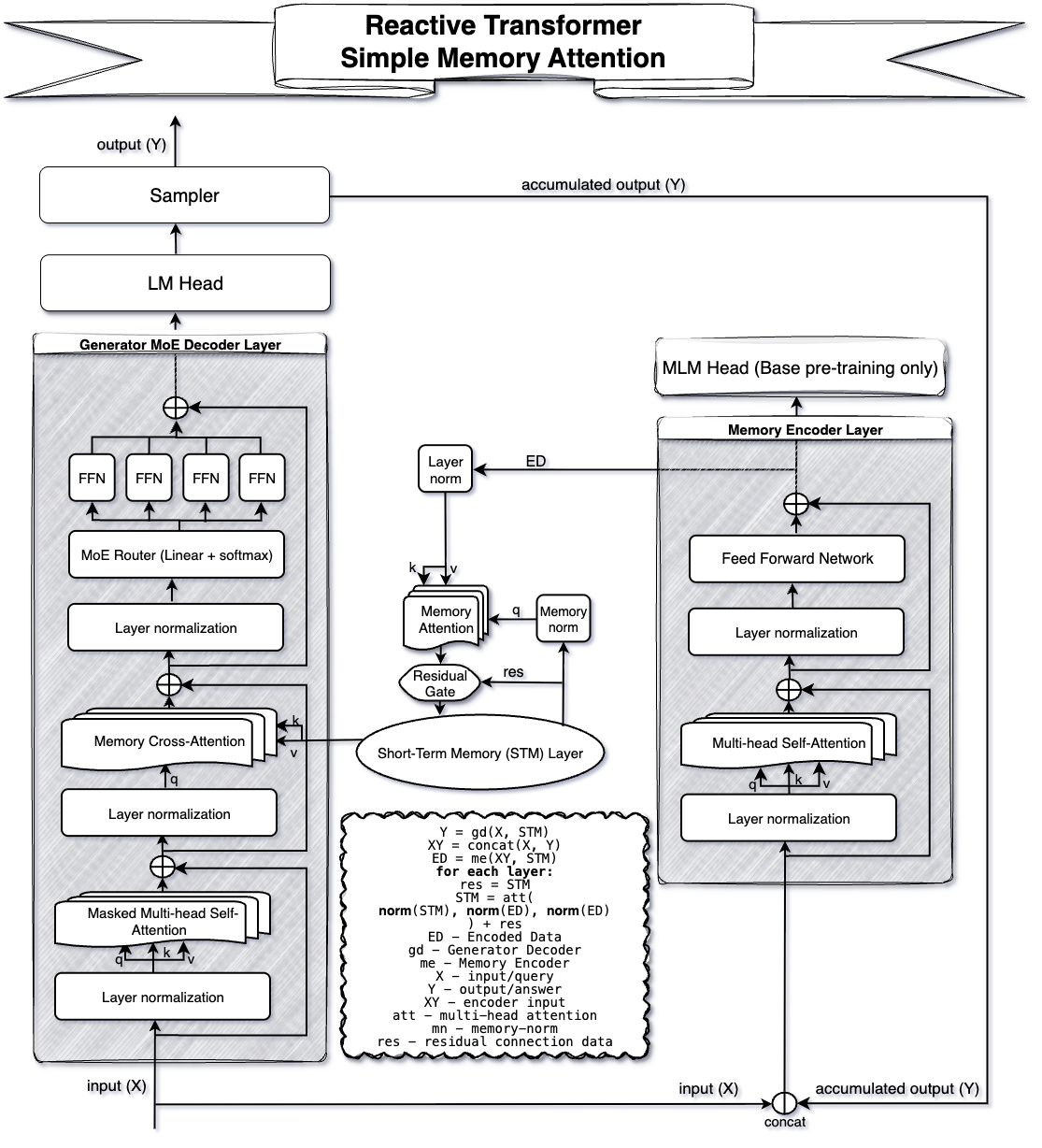}
    \caption{Simple Memory Attention variant}
    \label{fig:placeholder}
\end{figure}

\begin{figure}
    \centering
    \includegraphics[width=1\linewidth]{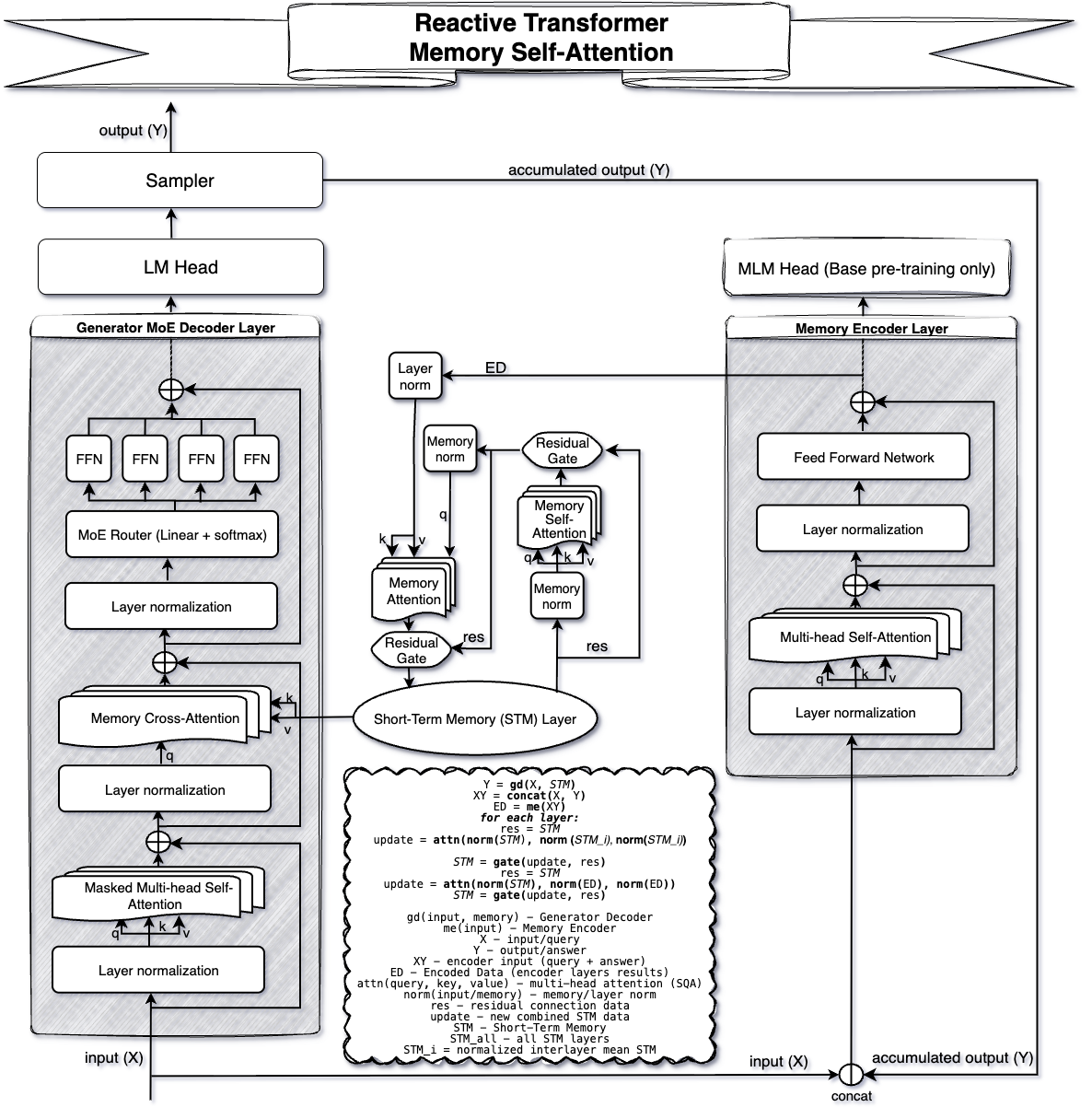}
    \caption{Memory Self-Attention variant}
    \label{fig:placeholder}
\end{figure}

\begin{figure}
    \centering
    \includegraphics[width=1\linewidth]{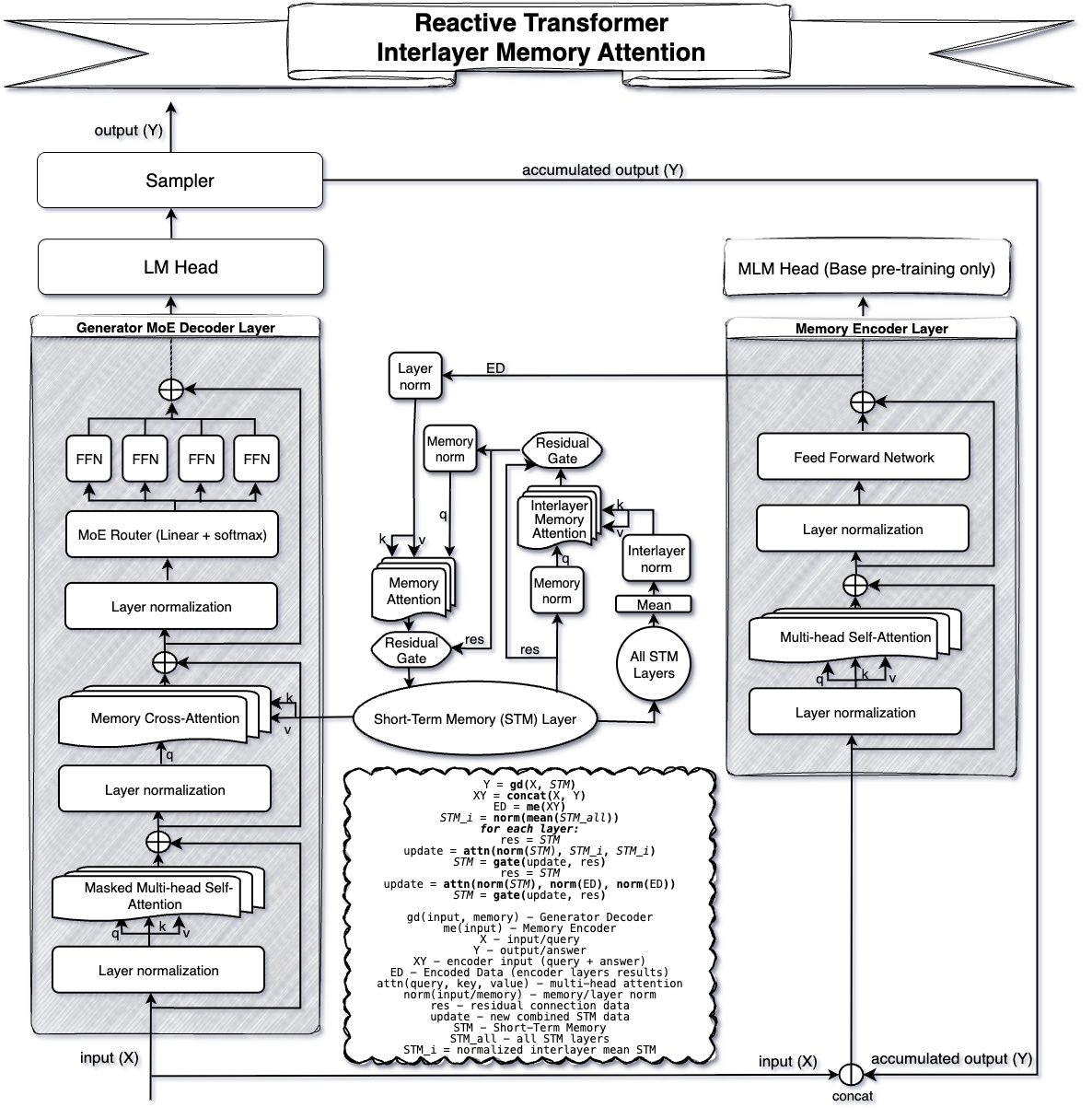}
    \caption{Interlayer Memory Attention variant}
    \label{fig:placeholder}
\end{figure}

\begin{figure}
    \centering
    \includegraphics[width=1\linewidth]{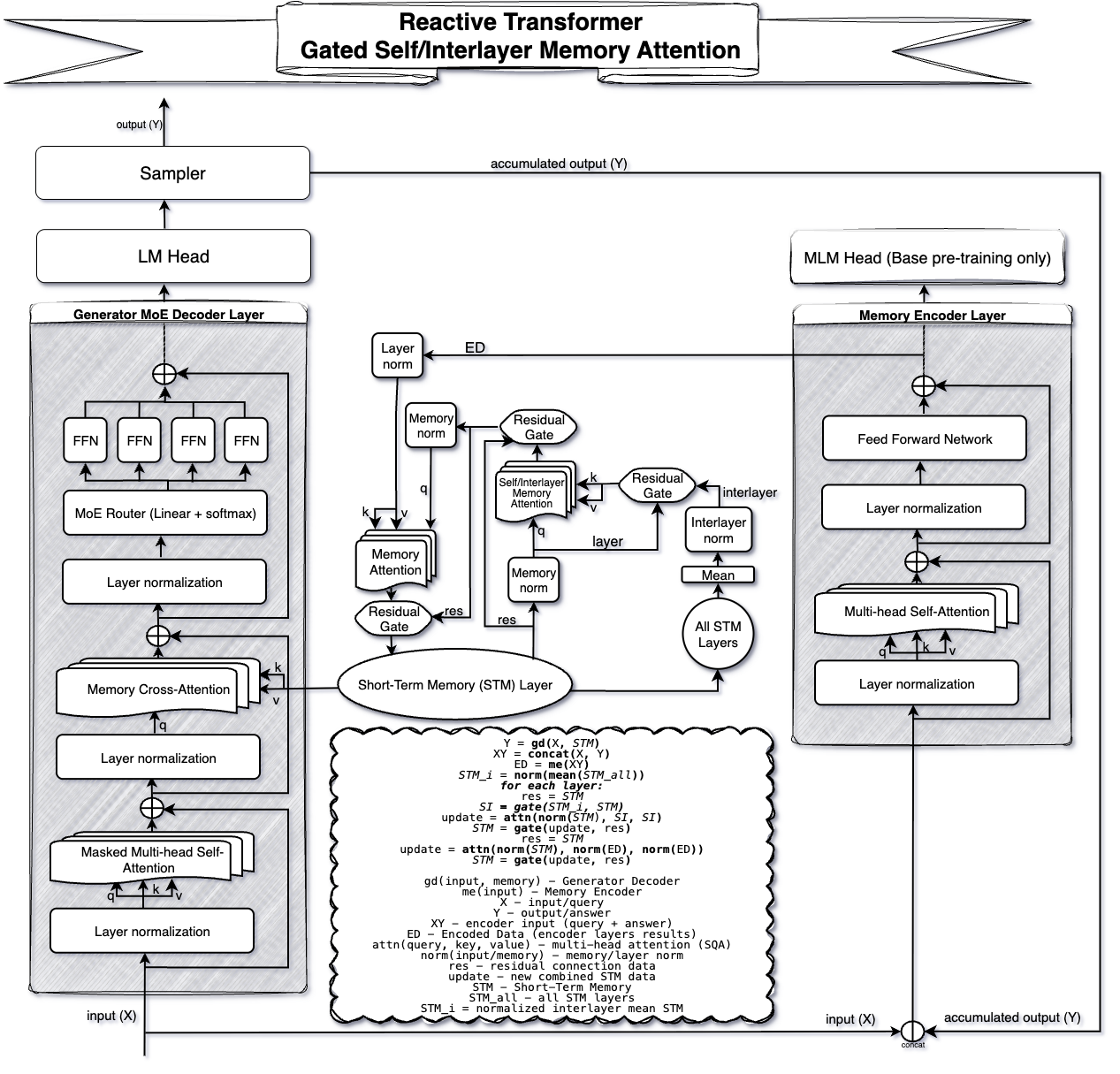}
    \caption{Gated Self/Interlayer Memory Attention variant}
    \label{fig:placeholder}
\end{figure}

\subsection{Architectural Components}

\textbf{Generator-Decoder:} The decoder is responsible for autoregressive text generation. Each of its layers follows a standard pre-norm structure but includes an additional \textbf{Memory Cross-Attention} sub-layer. The sequence of operations is: Masked Self-Attention, Memory Cross-Attention, and a Feed-Forward Network (FFN). To manage parameter counts effectively, the decoder's FFNs are implemented as Mixture-of-Experts (MoE) layers, allowing the decoder to have a much larger capacity than the encoder while maintaining a similar number of layers and hidden dimensions.

\textbf{Memory Encoder:} The encoder's sole purpose is to create a condensed representation of the completed interaction. It processes the concatenated query and response $concat(X, Y)$ through a series of standard encoder layers (Self-Attention and a dense FFN) to produce the hidden states that form the Encoded Data ($ED_t$).

\textbf{Memory Attention Network:} Memory attention is responsible for memory updates, based on the results of encoder - Encoded Data ($ED_t$). It has multiple variants with different attention layers configuration and could use \textbf{Memory Self-Attention} or \textbf{Interlayer Memory Attention} (or combination of both) to prepare $STM_{\text{t - 1}}$ for the final updates, when $STM_{\text{t - 1}}$ is combined with $ED_t$ to produce final $STM_{\text{t}}$. \textbf{Residual Gates} after each memory attention layer additionally decide how much the current and previous data should be used in the final updated state.

\textbf{Shared Embeddings:} The encoder, decoder, and memory system operate in a unified vector space. To facilitate this and reduce parameters, all components share a single token embedding layer.

\subsection{The Attention-Based Memory System (ABMS)}

\begin{figure}
    \centering
    \includegraphics[width=1\linewidth]{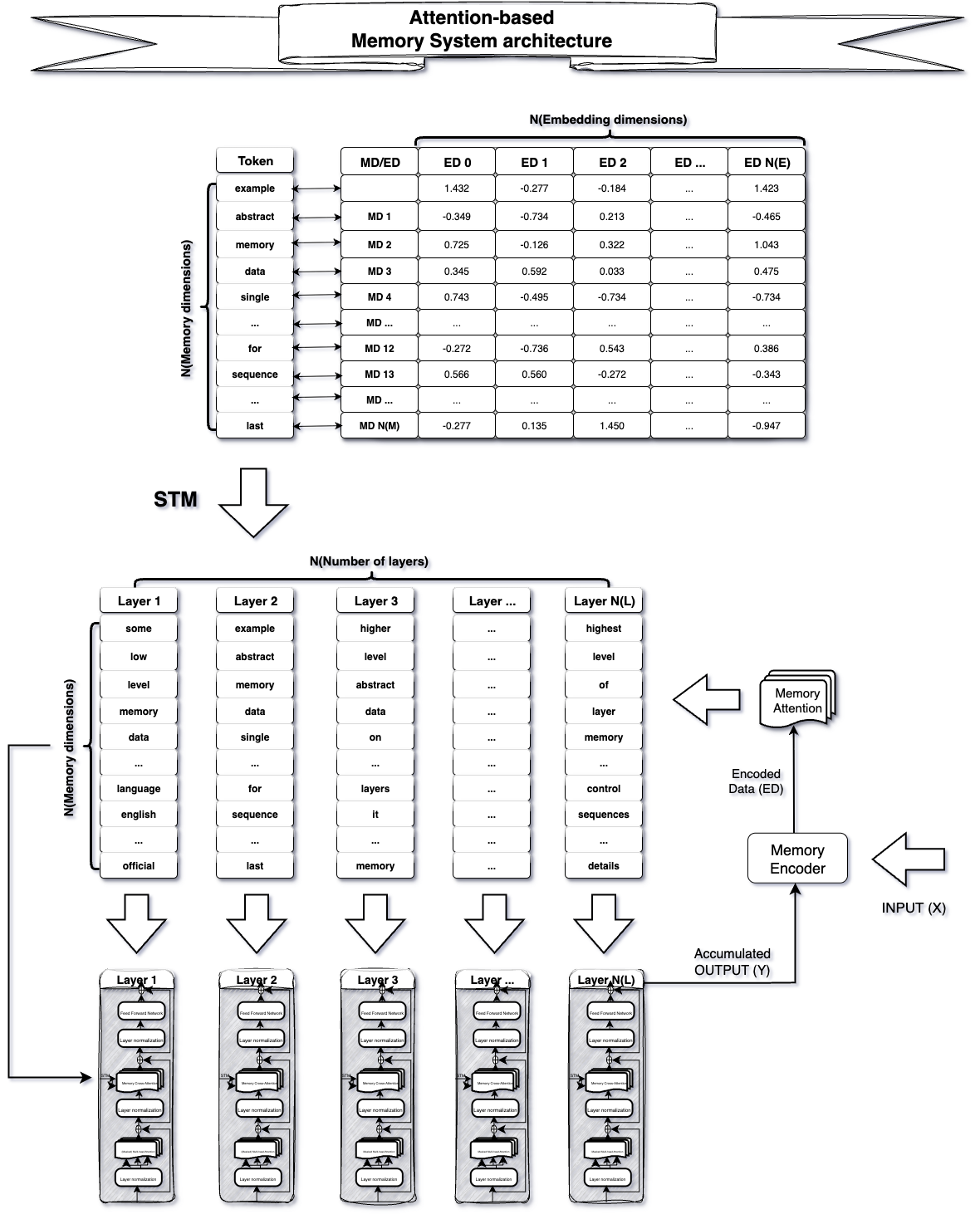}
    \caption{Attention-Based Memory System Architecture}
    \label{fig:placeholder}
\end{figure}

The core innovation of RxT lies in its integrated memory system. The STM is not a sequence of past tokens but a collection of fixed-size, learnable vectors (memory slots), organized into layers corresponding to each layer of the encoder and decoder.

\textbf{Memory Read (Memory Cross-Attention):} During generation, the decoder needs to access the conversational context stored in the STM. This is achieved via Memory Cross-Attention. In this operation, the hidden states of the decoder's input sequence act as the \textit{Queries} ($Q$), while the memory slots from the corresponding STM layer act as the \textit{Keys} ($K$) and \textit{Values} ($V$).
\begin{equation}
    \text{RetrievedContext} = \text{Attention}(Q=H_{\text{dec}}, K=STM_{t-1}, V=STM_{t-1})
    \label{eq:mem_read}
\end{equation}
Crucially, since the STM slots do not have an inherent sequential order, positional encodings (e.g., RoPE) are applied only to the queries ($H_{\text{dec}}$) and not to the keys and values from memory.

\textbf{Memory Write (Memory Attention):} The memory update process is conceptually the inverse of the read operation. Here, the memory slots from the previous state $STM_{t-1}$ act as the \textit{Queries} ($Q$), while the Encoded Data $ED_t$ from the Memory Encoder provides the \textit{Keys} ($K$) and \textit{Values} ($V$). This allows each memory slot to actively seek out and integrate relevant information from the latest interaction.
\begin{equation}
    \text{Update} = \text{Attention}(Q=STM_{t-1}, K=ED_t, V=ED_t)
    \label{eq:mem_write}
\end{equation}
The final memory state is then computed via a residual connection, typically controlled by a gate.

\subsection{Memory Attention Variants}

\begin{figure}
    \centering
    \includegraphics[width=0.75\linewidth]{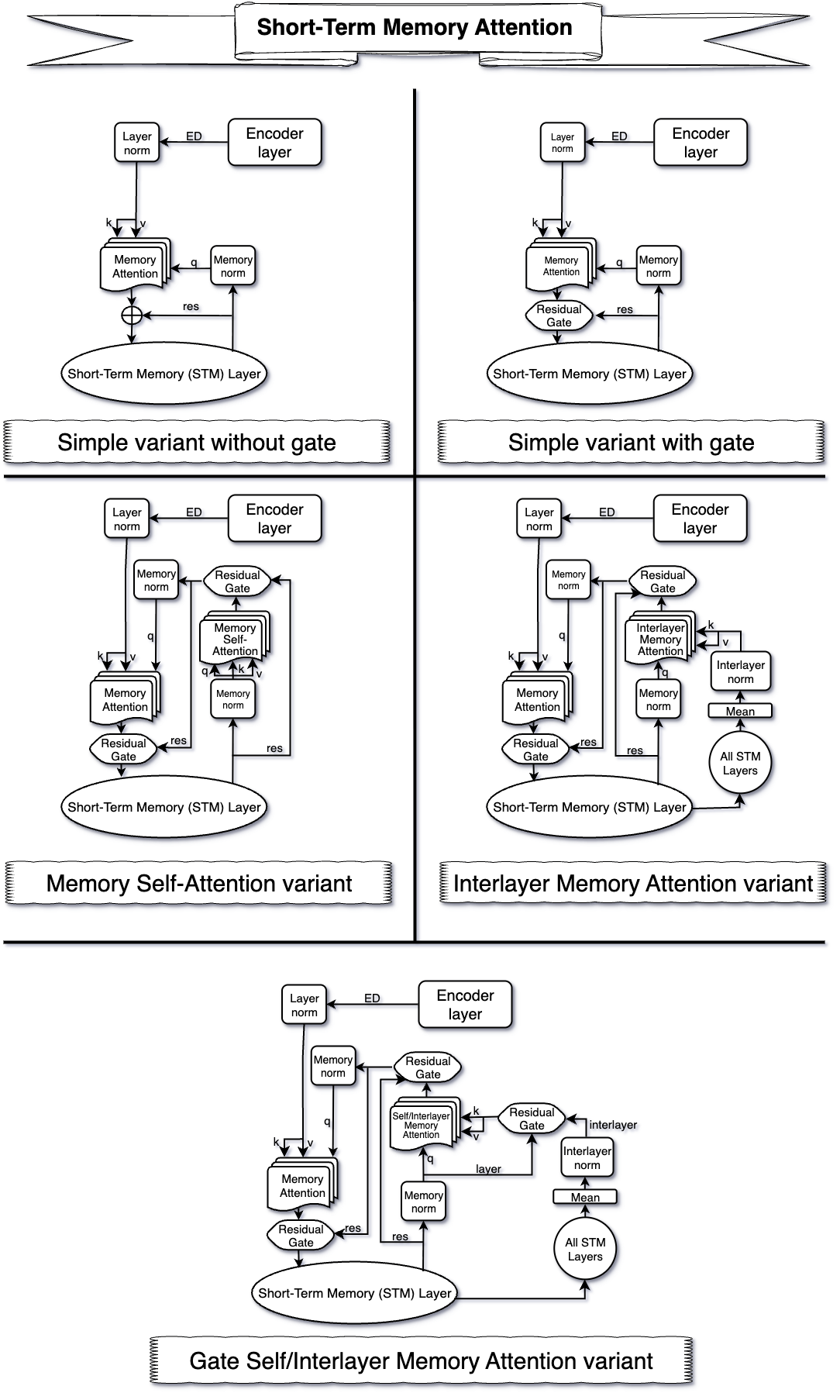}
    \caption{Memory Attention variants}
    \label{fig:placeholder}
\end{figure}

To provide flexibility in how memory is consolidated, we propose several variants for the Memory Attention network, as illustrated in Figures.

\begin{itemize}
    \item \textbf{Simple Memory Attention:} The most direct approach, where the STM queries the Encoded Data as described above.
    \item \textbf{Memory Self-Attention:} An additional self-attention step is introduced where memory slots attend to each other ($Q, K, V$ all from $STM_{t-1}$) before attending to the Encoded Data. This allows the model to reorganize and consolidate information within the memory itself.
    \item \textbf{Interlayer Memory Attention:} To facilitate communication across different levels of abstraction, memory slots in a given layer can also attend to an aggregated representation (e.g., the mean) of all other STM layers. This can help reduce information redundancy.
    \item \textbf{Gated Variants:} The flow of information from the self-attention or interlayer attention steps can be controlled by an additional Residual Gate, allowing the model to learn how much internal consolidation is needed.
\end{itemize}

\subsection{Residual Gates}

\begin{figure}
    \centering
    \includegraphics[width=0.75\linewidth]{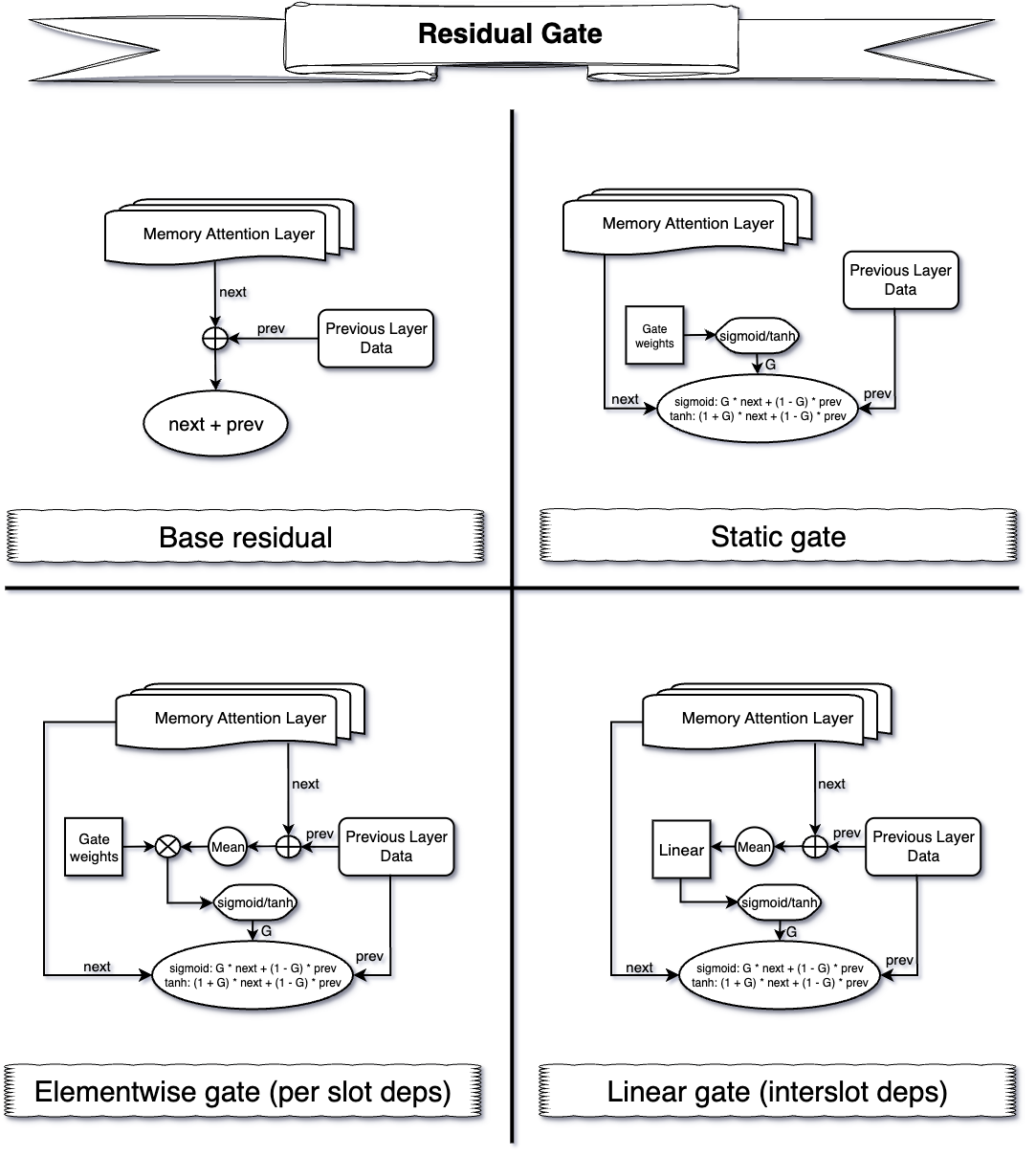}
    \caption{Residual Gate variants}
    \label{fig:placeholder}
\end{figure}

To prevent catastrophic forgetting and control the plasticity of the memory, the residual update is managed by a \textbf{Residual Gate}. Instead of a simple addition ($STM_t = STM_{t-1} + \text{Update}$), the gate computes a dynamic interpolation:
\begin{equation}
    STM_t = (1 - G) \odot STM_{t-1} + G \odot \text{Update}
    \label{eq:res_gate}
\end{equation}
where $G$ is a gating vector computed by a small network, typically using a sigmoid activation function. The sigmoid ensures that the update is a weighted average, which empirically prevents the magnitude of the memory vectors from exploding over many interactions and provides more stable training. The gate computation can be dynamic (dependent on the input states) or static (a learned constant). Alternatively, \textbf{tanh} activation could be used or the gating could be completely skipped, but it requires much stronger regularization of memory updates in training algorithms. The gate result with \textbf{tanh} activation is calculated with:
\begin{equation}
    STM_t = (1 - G) \odot STM_{t-1} + (1 + G) \odot \text{Update}
    \label{eq:res_gate_tanh}
\end{equation}

\subsection{Comparison with Stateful Memory-Augmented Transformers}
Before \textbf{Reactive Transformer (RxT)}, there were the attempts to provide stateful conversation processing with linear cost scaling. Furthermore, those stateful models achieved better results than comparable stateless models, especially in long multi-turn conversations. It makes the decision to proceed with stateless LLMs even more irrational. However, there are notable differences between RxT and Stateful MAT, that made RxT fundamental paradigm shift instead of being just an evolution.

\subsubsection{Compatibility with event-driven paradigm}
\textbf{Event-Driven AI} paradigm made for stateful real-time processing is based on two types of events handled by the conversational language models: \textbf{query/input event} (current user's query) and \textbf{response/output event} (model's answer). Process of generating \textbf{response/output event} in reaction on \textbf{query/input event} is defined as \textbf{interaction}. Synchronous memory update in encoder-first stateful MATs requires previous model's answer to be passed with current query as an input, to preserve all dialog history context. It breaks the fundamental \textbf{Event-Driven AI} rule, by processing events from two conversation steps at once (and leads to higher encoding cost and latency). In the opposite, RxT's input is always only current user's query.
That difference has also practical implications:
\begin{itemize}
    \item \textbf{Stateful MAT:} Memory is accessed and updated (in encoder), basing on previous answer and current query.
    \item \textbf{RxT:} Memory is accessed (in decoder) and updated (by encoder and memory attention), basing on current interaction: query and answer. That's far more natural and expressive, because model knows for which query the answer were generated.
\end{itemize}

\begin{figure}
    \centering
    \includegraphics[width=1\linewidth]{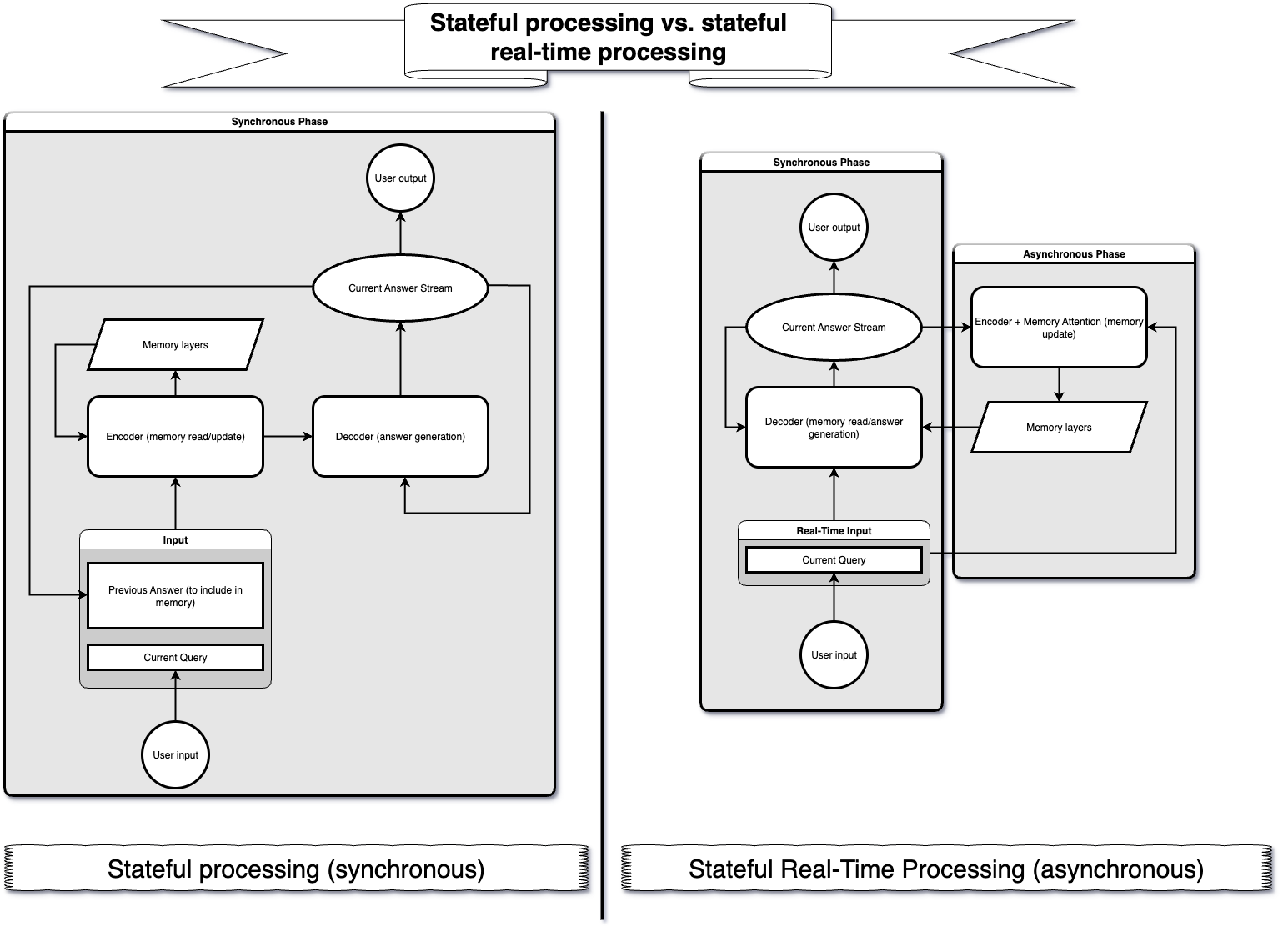}
    \caption{Diagrams of stateful and stateful real-time processing}
    \label{fig:placeholder}
\end{figure}

\subsubsection{Single Responsibility Principle}
In encoder-first stateful MATs, the encoder is responsible for memory access and update, while the decoder is using combined memory and input in its cross-attention. In Reactive Transformer, components responsibility is separated, leading to easier tasks for each component and better expressiveness in result:
\begin{itemize}
    \item \textbf{Decoder's self-attention} is responsible for current interaction inter-sequence dependencies - both query and generated answer are treated as single interaction input to self-attention. It's closer to LLMs text completion, but without the long conversation history.
    \item \textbf{Decoder's memory cross-attention} is responsible for memory access for processed query and generated answer. It contains only the information from previous steps, not polluted by input data (that's handled by self-attention).
    \item \textbf{Encoder} has only self-attention and responsibility to transform the current interaction into memory vector spaces. That's a simple, full sequence processing task, without any memory access and update - that's why encoder could be a lot smaller (dense vs. MoE decoder) to balance the overall parameters count
    \item \textbf{Memory Attention network} is responsible for memory updates, based on current interaction and previous memory state. Due to the separation from the encoder, it could have multiple expressive variants.
\end{itemize}

\subsubsection{Design choices for architecture and training pipelines}
Stateful MATs and Reactive Transformer were designed with different goals in mind and started from different baselines.

\textbf{Stateful MATs} were created to extend the encoder-decoder transformer with memory and are made for stateful sequence to sequence processing. They are using the same decoder as baseline seq-to-seq models and modifying only the encoder. The advantage of this approach is only slightly modified training pipeline.

On the other hand, \textbf{Reactive Transformer} was designed to extend the decoder-only LLMs with stateful processing and memory system, and replace them in conversational tasks. It changes the data-driven sequence completion into event-driven interaction completion, instead of using sequence to sequence processing. RxT was created from scratch, without any baseline and require completely re-designed training pipeline to handle its asynchronous nature. It's natively trained for conversational tasks and rather cannot be used for single-step generative tasks.

\section{Inference and Computational Analysis}

The architecture of \textbf{Reactive Transformer (RxT)} leads to a highly efficient inference process and fundamentally different scaling properties compared to traditional LLMs, and asynchronous memory update eliminates Stateful MAT latency overhead. 

\subsection{Inference Process}

A single interaction in RxT consists of two phases:

\begin{enumerate}
    \item \textbf{Prompt Processing \& Autoregressive Generation:}
    \begin{itemize}
        \item The input query $X_t$ is processed in a single forward pass through the decoder to populate a self-attention KV cache. This is the "prompt phase".
        \item The decoder then generates the response $Y_t$ token by token. For each token, it attends to the self-attention KV cache (containing the query and previously generated response tokens) and the memory cross-attention KV cache.
        \item \textbf{Optimization:} Since the memory state $STM_{t-1}$ is static throughout the generation of a single response, its Key and Value projections for the Memory Cross-Attention mechanism can be \textbf{fully pre-cached} before the prompt phase. This eliminates redundant projections for every generated token, significantly accelerating the generation process.
    \end{itemize}
    \item \textbf{Asynchronous Memory Update:}
    \begin{itemize}
        \item Once generation is complete, the concatenated interaction $concat(X, Y)$ is processed by the Memory Encoder and Memory Attention network to compute $STM_t$. This happens "in the background" and does not add to the user-perceived latency.
    \end{itemize}
\end{enumerate}

\subsection{Computational Cost Analysis}

To analyze the computational cost, let $N$ be the number of interactions in the conversation history, $T_{\text{query}}$ be the length of the current query, $T_{\text{answer}}$ be the length of the generated answer, and $T$ be the average length of a full interaction ($T \approx T_{\text{query}} + T_{\text{answer}}$).

\textbf{Standard Transformer LLM:}
At interaction $N+1$, the model must first process a prompt of length $L_{\text{prompt}} \approx (N \cdot T) + T_{\text{query}}$. Subsequently, it generates $T_{\text{answer}}$ tokens autoregressively.

\begin{itemize}
    \item \textbf{Single Interaction Cost (Computational):} The total cost is a sum of two phases:
    \begin{enumerate}
        \item \textbf{Prompt Processing:} This phase is parallelizable but has a cost quadratic to the input length: $O(L_{\text{prompt}}^2) \approx O((N \cdot T + T_{\text{query}})^2)$.
        \item \textbf{Token Generation:} This phase is sequential. To generate each of the $T_{\text{answer}}$ tokens, the model must perform an attention operation over a KV cache that contains the entire preceding conversation history. The cost to generate a single token is therefore linear with the cache length, which is approximately $L_{\text{prompt}}$. The total generation cost is thus $O(L_{\text{prompt}} + T_{\text{answer}}) \approx O((N + 1) \cdot T)$.
    \end{enumerate}
    The combined cost is dominated by these two terms: $O((N \cdot T + T_{\text{query}})^2 + (N + 1) \cdot T)$. As $N$ grows, both prompt processing and per-token generation become prohibitively expensive.
    \item \textbf{Total Conversation Cost (User-Facing):} From a user-cost perspective (e.g., API pricing based on token count), the total number of tokens processed across $N$ interactions scales quadratically. The total number of tokens is approximately $\sum_{k=1}^{N} (k \cdot T)$, which results in a cost of $O(N^2 \cdot T)$.
\end{itemize}

\textbf{Reactive Transformer (RxT):}
At any interaction, the model's decoder processes a query of length $T_{\text{query}}$ and generates a response of length $T_{\text{answer}}$. The memory update cost is handled asynchronously.

\begin{itemize}
    \item \textbf{Single Interaction Cost (Computational):}
    \begin{enumerate}
        \item \textbf{Prompt Processing:} The cost is quadratic only to the current, short query: $O(T_{\text{query}}^2 + T_{\text{query}} \cdot S_{\text{mem}})$.
        \item \textbf{Token Generation:} In stark contrast to standard LLMs, the KV cache for self-attention only contains tokens from the \textit{current} interaction, and the memory cross-attention cache is of a fixed size. Therefore, the computational cost to generate each subsequent token is not only low but, critically, \textbf{constant} with respect to the overall conversation length $N$. The total generation cost is approximately $O(T_{\text{query}} + T_{\text{answer}} + S_{\text{mem}})$.
        \item \textbf{Asynchronous memory update:} RxT has additional memory update step after all response token generation. It scales quadratically with short-term memory size $O(S_{\text{mem}}^2)$, but is not influencing the user's latency.
    \end{enumerate}
    The combined user-perceived cost is $O(T_{\text{query}}^2 + T_{\text{query}} \cdot S_{\text{mem}} + T_{\text{query}} + T_{\text{answer}} + S_{\text{mem}})$, and additional update cost $O(S_{\text{mem}}^2)$, the values independent of $N$.
    \item \textbf{Total Conversation Cost (User-Facing):} The model processes a roughly constant number of tokens, $T$, for each of the $N$ interactions. Therefore, the total user-facing cost scales linearly: $O(N \cdot T)$.
\end{itemize}

This analysis reveals a dual advantage for RxT: not only is the initial prompt processing exponentially cheaper for long conversations, but the \textbf{per-token generation speed remains constant}, whereas for standard LLMs, it degrades linearly with the length of the dialogue. Table \ref{tab:costs} summarizes this comparison.

\begin{table}[h]
  \caption{Asymptotic comparison of computational costs for a single message of T tokens in conversation of N turns.}
  \label{tab:costs}
  \centering
  \begin{tabular}{llll}\toprule
    
    \textbf{Model Architecture} & \textbf{Prompt phase cost}&   \textbf{Generate phase}& \textbf{Asynchronous phase}\\
 & \textbf{(Computational, turn N)}& \textbf{(Computational, turn N)}&\textbf{(Computational, slots S)}\\ \midrule
    LLM& $O((N \cdot T)^2)$&   $O((N \cdot T) + T_{\text{answer}})$& -\\
 & $O(T \cdot (N \cdot T))$***& &\\
 LLM with RAG& $O((T_{\text{mem}} + N \cdot T)^2)$& $O((T_{\text{mem}} + N \cdot T) + T_{\text{answer}})$&-\\
    SSM / Linear Attn& $O(N \cdot T)$ &   $O((N \cdot T) + T_{\text{answer}})$& -\\
    Synchronous MAT& $O(T^2 + S_{\text{mem}}^2)$&   $O(T_{\text{answer}} + S_{\text{mem}})$& -\\
 & $O(S_{\text{mem}}^2)$*& &\\
    \textbf{RxLM (RxT)}& \textbf{$O(T_{\text{query}}^2 + T_{\text{query}} \cdot S_{\text{mem}})$}&   \textbf{$O(T_{\text{query}} + T_{\text{answer}} + S_{\text{mem}})$}& $O(S_{\text{mem}}^2)$**\\
    
 & $O(T_{\text{query}} \cdot S_{\text{mem}})$*& &\\ \bottomrule 
  \end{tabular}
\end{table}

\begin{table}[h]
  \caption{Comparison of end user costs for a conversation of N turns.}
  \label{tab:costs}
  \centering
  \begin{tabular}{llll}
    \toprule
    \textbf{Model Architecture} & \textbf{Single Interaction Cost} & \textbf{Total User-Facing Cost} & \textbf{Scaling vs.} \\
    & \textbf{(Tokens)}& \textbf{(Tokens)} & \textbf{Interactions} \\
    \midrule
    LLM& $O(N \cdot T)$& $O(N^2 \cdot T)$ & \textbf{Quadratic} \\
    SSM / Linear Attn& $O(N \cdot T)$ & $O(N^2 \cdot T)$ & \textbf{Quadratic} \\
    Synchronous MAT& $O(T)$& $O(N \cdot T)$ & \textbf{Linear}\\
    \textbf{Reactive Transformer (RxT)} & \textbf{$O(T)$}& \textbf{$O(N \cdot T)$} & \textbf{Linear} \\
    \bottomrule
  \end{tabular}
\end{table}

* $S_{\text{mem}} > T_{\text{query}}$, so in \textbf{Synchronous MAT} prompt phase $S_{\text{mem}}^2$ is the dominant factor, while in \textbf{RxT} $T_{\text{query}} \cdot S_{\text{mem}}$ dominates.

**  Thanks to the asynchronous update phase, \textbf{RxT} includes answers in memory, so it has full dialog context without delays for the user.

*** With persistent KV-cache

\section{Supervised Training Curriculum}

The architectural design of the Reactive Transformer, with its asynchronous, multi-component nature, presents unique training challenges that cannot be addressed by a naive, end-to-end optimization strategy. Such an approach is prone to instability and convergence failure due to the complex interplay between the generator, encoder, and the non-interpretable memory state. To overcome these hurdles, a carefully designed, multi-stage supervised training curriculum is proposed. This curriculum acts as a scaffolding process, systematically pre-training and integrating each component to solve predictable failure modes before unifying the entire system. The objective is to establish a shared semantic space, pre-condition the memory network for its abstract task, and finally, teach the full, memory-dependent interaction cycle.

\subsection{Overview of the Training Pipeline}

The supervised training pipeline consists of four distinct stages, designed to incrementally build the model's capabilities. The overarching strategy is to first establish a robust foundation for the language components, then independently prepare the memory system, and finally, integrate them to learn the complete, stateful conversational flow. This structured approach is essential for solving the "cold start" problem inherent in two key areas: the interaction between the decoder and the memory system, and the memory update mechanism itself. By addressing these challenges systematically, the curriculum prepares the model for the subsequent, more advanced reinforcement learning phases. The four supervised stages are:
\begin{enumerate}
    \item \textbf{Joint Language Model Pre-Training:} Co-trains the Generator-Decoder and Memory Encoder on a large text corpus to learn fundamental language representations and align their vector spaces.
    \item \textbf{Joint Interaction Supervised Fine-Tuning (SFT):} Adapts the pre-trained language components to the specific format of conversational interactions using a dataset of query-response pairs.
    \item \textbf{Self-Supervised Memory Attention Pre-Training:} Trains the Memory Attention network on a proxy task to produce semantically coherent outputs, addressing the lack of direct ground-truth labels for memory states.
    \item \textbf{Supervised Memory-Aware Training:} Unifies all pre-trained components and trains the full model on multi-step dialogues, teaching the decoder to leverage the accumulated memory state for maintaining context.
\end{enumerate}

\subsection{Stages 1 \& 2: Joint Language Model Pre-Training and Fine-Tuning}

The initial stages are focused on co-training the Generator-Decoder and Memory Encoder to establish a shared semantic foundation. The model is treated analogously to a standard encoder-decoder transformer, employing a dual-objective function.

The training algorithm proceeds as follows: An input sequence $S$ is duplicated. One copy is processed autoregressively by the decoder, while the other is randomly masked to create $S_{\text{mask}}$ for the encoder. The Memory Encoder processes $S_{\text{mask}}$, and its final hidden states are passed to a dedicated Masked Language Modeling (MLM) head to compute the MLM loss, $\mathcal{L}_{\text{MLM}}$.

Concurrently, the hidden states from each layer of the encoder, $ED = \{ed_1, ed_2,..., ed_L\}$, are detached from the computation graph. This crucial step prevents gradients from the decoder from flowing back into the encoder, effectively treating the encoder's output as a fixed target for the decoder. To improve generalization and prevent the decoder from becoming overly reliant on this perfect context, a small amount of random noise, $\epsilon$, is added: $ED' = ED + \epsilon$. These noisy states serve as the Key and Value inputs for the decoder's Memory Cross-Attention layers. The decoder then processes the original, unmasked sequence $S$ autoregressively, conditioned on the context from $ED'$, and the standard autoregressive cross-entropy loss, $\mathcal{L}_{\text{AR}}$, is computed. The total loss for the joint training is a weighted sum:
\begin{equation}
    \mathcal{L}_{\text{Joint}} = \alpha \mathcal{L}_{\text{AR}} + \beta \mathcal{L}_{\text{MLM}}
\end{equation}
where $\alpha$ and $\beta$ are hyperparameters balancing the two objectives. This "teacher forcing" approach rapidly bootstraps the connection between the encoder and decoder. The addition of noise acts as a vital regularization technique, mitigating the primary risk of this stage: the decoder developing a weak self-attention mechanism by depending too heavily on the "cheated" context provided by the encoder.

The Supervised Fine-Tuning (SFT) stage follows the exact same algorithm but shifts the data distribution from a general text corpus to a dataset of structured conversational turns, typically formatted with special tokens (e.g., `[Query]... [Answer]`). This adapts the model to the specific turn-taking format of dialogue.

\subsection{Stage 3: Self-Supervised Memory Attention Pre-Training}

The central challenge in training the Memory Attention network is that its target output—the updated memory state $STM_t$—is a high-dimensional, non-interpretable tensor for which no human-generated labels exist. To circumvent this, a self-supervised proxy task is employed. The objective is to train the network to produce a plausible combination of the previous memory state and the new information from the current interaction.

The algorithm is initialized with a previous memory state $STM_{t-1}$ (initially random noise) and the Encoded Data $ED_t$ from a pre-trained encoder. A pseudo-label, $STM_{\text{target}}$, is generated via a dynamic weighted average:
\begin{equation}
    STM_{\text{target}} = (1 - w_t) \cdot STM_{t-1} + w_t \cdot ED_t
\end{equation}
The weighting factor $w_t$ is annealed over a sequence of interactions within a curriculum. For the first interaction, $w_t$ is high (e.g., 0.9) to prioritize incorporating the new information. For subsequent interactions, $w_t$ is progressively decreased, encouraging retention and integration. The Memory Attention network then computes the actual updated state, $STM_t = \text{MemAttn}(STM_{t-1}, ED_t)$. The loss function is the negative cosine similarity between the predicted and target states, which encourages semantic alignment without enforcing an exact match:
\begin{equation}
    \mathcal{L}_{\text{Mem}} = - \text{cosine\_similarity}(STM_t, STM_{\text{target}})
\end{equation}
This training stage is a critical piece of the curriculum's scaffolding. A randomly initialized Memory Attention network would output vectors that are effectively noise. If this noisy output were fed directly to the decoder in the next stage, it would act as a powerful distractor, corrupting the learning signal. The decoder would likely learn to ignore its memory cross-attention layers entirely, defeating the architecture's purpose. Empirical evidence confirms this failure mode, with initial reinforcement learning rewards dropping to near-zero without this pre-training step. Therefore, the primary function of this stage is not to create a perfect memory system, but to pre-condition the network to produce outputs that are semantically coherent and reside within the same vector space as the other components, thus solving the cold start problem and enabling subsequent training stages to succeed.

\subsection{Stage 4: Supervised Memory-Aware Training}

The final supervised stage unifies all pre-trained components to train the model on its intended, event-driven operational cycle. This is the first point at which the decoder learns to rely on a meaningful, accumulated memory state from genuinely past interactions, rather than the "cheated" context from the joint training stages.

The algorithm uses a curriculum of multi-step dialogues, $\{I_1, I_2,..., I_N\}$.
\begin{enumerate}
    \item The memory state $STM_0$ is initialized with random noise (e.g., from a normal distribution). The model processes the first interaction $I_1 = (X_1, Y_1)$, with the decoder conditioned on $STM_0$. The autoregressive loss $\mathcal{L}_1$ is computed. This step explicitly trains the model to handle the beginning of a conversation from a blank state.
    \item For each subsequent step $t$ from 1 to $N-1$:
    \begin{enumerate}
        \item The completed interaction $I_t$ is encoded using the Memory Encoder to produce $ED_t$.
        \item The memory state is updated using the pre-trained Memory Attention network: $STM_t = \text{MemAttn}(STM_{t-1}, ED_t)$.
        \item The decoder processes the next interaction's query $X_{t+1}$ and generates the response $Y_{t+1}$, conditioned on the newly computed memory state $STM_t$.
        \item The autoregressive loss $\mathcal{L}_{t+1}$ is computed for this interaction.
    \end{enumerate}
    \item The total loss is the sum of losses from all steps, and backpropagation is performed. To stabilize training, the parameters of the encoder and memory attention network may be initially frozen and then gradually unfrozen.
\end{enumerate}
This stage directly optimizes the model's ability to maintain conversational coherence by forcing the decoder to extract relevant context from the dynamically updated STM. Upon completion, the model possesses a partially functional memory system and is fully prepared for refinement via reinforcement learning.

\section{Experiments and Results}

To validate the architectural claims and the effectiveness of the training curriculum, a series of experiments were conducted. The primary research questions were: (1) Does the RxT architecture outperform a standard decoder-only LLM of comparable size on multi-turn conversational tasks? (2) Do the performance benefits of RxT scale with model size? (3) Does the proposed training curriculum successfully enable the memory system to function effectively?

\subsection{Experimental Setup}

\textbf{Models:} Four RxT variants of increasing scale were trained, alongside a baseline model for comparison.
\begin{itemize}
    \item \textbf{RxT Variants:} RxT-Alpha Nano (12M parameters), RxT-Alpha Micro (26M), RxT-Alpha Mini (100M), and RxT-Alpha Synthetic (160M). The Nano model used Interlayer Memory Attention, while the larger variants used the more expressive Gated Self/Interlayer Memory Attention. Interlayer variants achieved better results than simple/self memory attention in supervised memory-aware training experiments, so we are using them as a default choice.
    \item \textbf{Baseline:} A 22M parameter decoder-only Transformer LLM was trained on the same data to provide a direct and fair comparison point for the smaller RxT models.
\end{itemize}

\textbf{Datasets:} All models were trained using the TinyStories \cite{eldan2023stories} dataset for general language pre-training \cite{patil2025stories}. Subsequently, custom multi-turn interaction datasets derived from TinyStories were used for fine-tuning and evaluation. These datasets, referred to as MRL Curriculum Datasets, consist of series of interconnected interactions designed to test context retention.

\textbf{Models architectures:} All RxT models, as well as reference stateless decoder-only model are using our Sparse Query Attention (SQA) \cite{filipek2025sqa} for all attention layers. It's computational efficiency is a perfect match for RxT Encoder and Memory Attention networks, because they are based on full sequence processing. All the models are also using Mixture-of-Experts in decoders. Baseline LLM is using exactly the same configuration for the fair comparison. We could compare the models with similar size GPT-1 \cite{radford2018} and GPT-2 \cite{brown2020} models or stateful MemBART \cite{wu2022stateful}, but those models are old and have outdated architectures, so we decided, that it will be unfair comparison - it's not hard to outperform models from previous generations. While stateful predecessors like MemBART  offer important context, our primary goal is to demonstrate the advantages of RxT's event-driven paradigm over the dominant stateless approach used in today's state-of-the-art systems

\textbf{Evaluation Metrics:} Performance was assessed using three metrics:
\begin{itemize}
    \item \textbf{Perplexity (PPL):} A standard measure of a language model's ability to predict a sequence of text. Lower values indicate better fluency.
    \item \textbf{Accuracy:} Standard next-token prediction accuracy.
    \item \textbf{MRL Reward Score:} A custom, composite metric designed to serve as a proxy for conversational quality, scaled to a 0-10 range. It is a weighted sum of BLEU score (for fluency), cosine similarity between the generated response and the ground-truth response (for immediate relevance), and previous interaction content, with cosine similarity between the generated response and the history of preceding ground-truth interactions (for long-term coherence). This reward score methodology will be deeply described in one of our next research papers for memory benchmark.
\end{itemize}

\subsection{Performance on Memory-Aware Language Modeling}

The results from the final Supervised Memory-Aware Training stage provide a clear quantitative measure of the models' language modeling capabilities. Table \ref{tab:ppl_results} summarizes the performance of all models on the multi-turn dialogue test set.

\begin{table}[h]
  \caption{Memory-Aware Training Performance on Multi-Step Dialogues.}
  \label{tab:ppl_results}
  \centering
  \begin{tabular}{lccc}
    \toprule
    \textbf{Model} & \textbf{Parameters} & \textbf{Perplexity (PPL)} & \textbf{Accuracy (\%)} \\
    \midrule
    LLM Baseline & 22M & 4.37 & 55 \\
    \midrule
    RxT-Alpha Nano & 12M & 2.74 & $\sim$81 \\
    RxT-Alpha Micro & 26M & 2.56 & $\sim$82 \\
    RxT-Alpha Micro (updated) & 26M & 2.31 & $\sim$82 \\
    RxT-Alpha Mini & 100M & 2.50 & $\sim$80 \\
    RxT-Alpha Synthetic & 160M & 2.18 & $\sim$82 \\
    \bottomrule
  \end{tabular}
\end{table}

\begin{figure}
    \centering
    \includegraphics[width=0.75\linewidth]{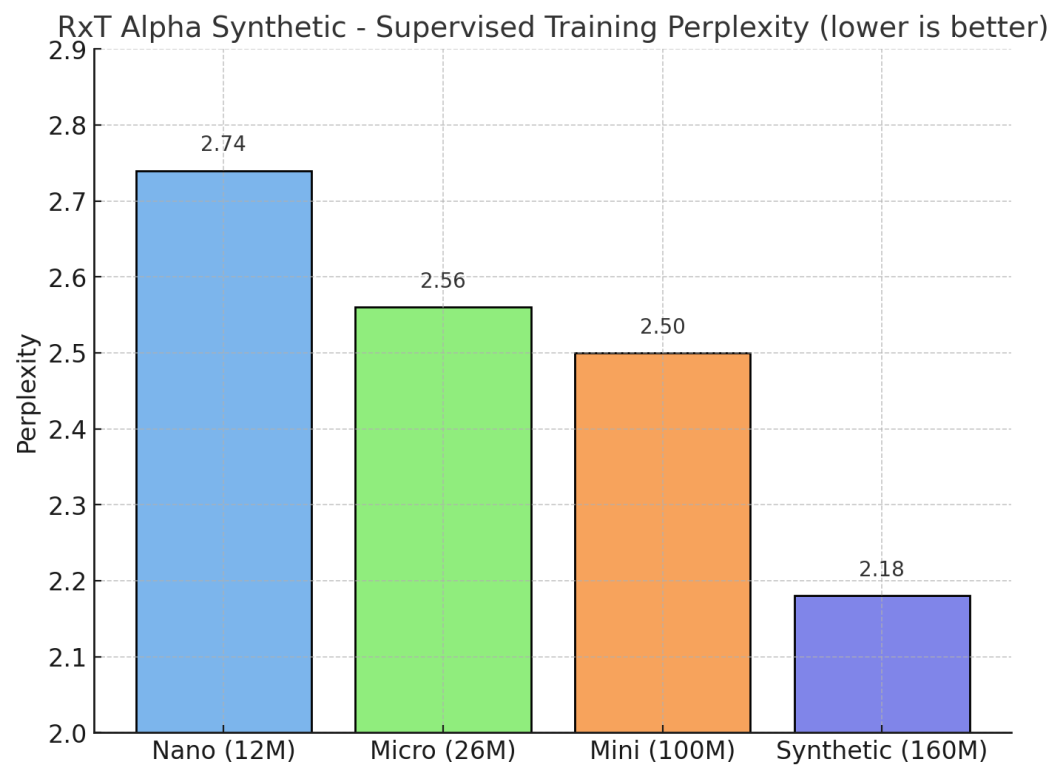}
    \caption{Supervised Dialog Perplexity of different scale models}
    \label{fig:placeholder}
\end{figure}

\begin{figure}
    \centering
    \includegraphics[width=0.5\linewidth]{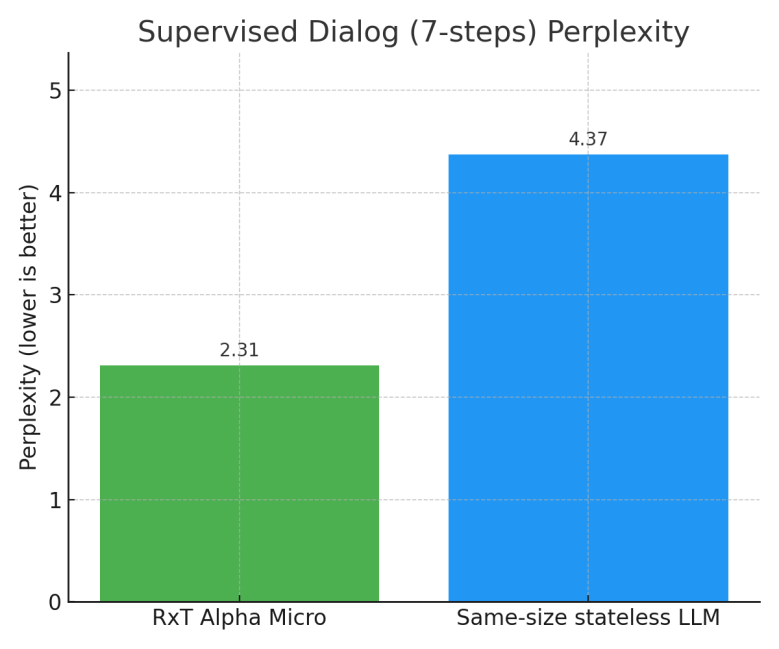}
    \caption{Supervised Dialog Perplexity compared to the same size stateless LLM}
    \label{fig:placeholder}
\end{figure}

\begin{figure}
    \centering
    \includegraphics[width=0.5\linewidth]{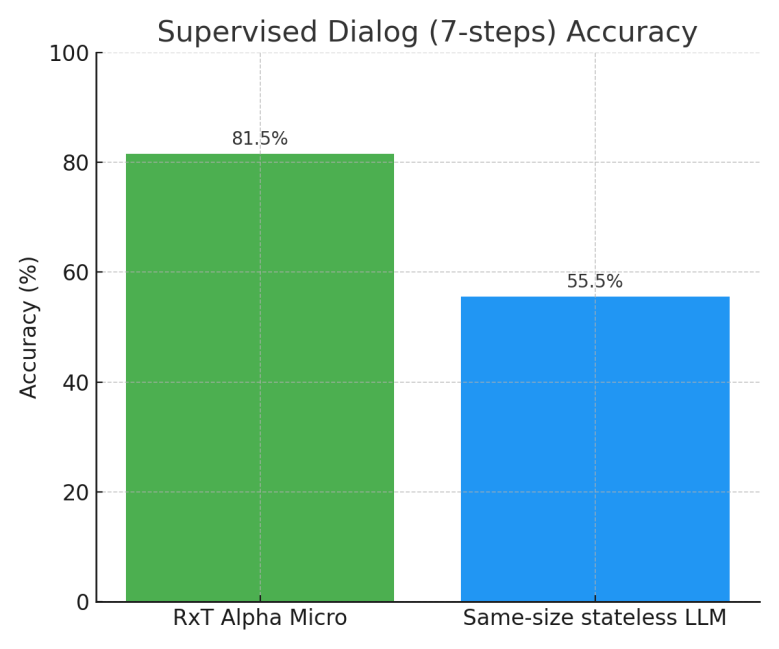}
    \caption{Supervised Dialog Accuracy compared to the same size stateless LLM}
    \label{fig:placeholder}
\end{figure}

The results demonstrate a stark performance gap. Even the smallest RxT model (Nano, 12M) significantly outperforms the larger 22M LLM baseline, achieving a perplexity of 2.74 compared to the baseline's 4.37. This trend holds across all scales, with every RxT variant showing superior fluency and predictive accuracy. The data also confirms that the RxT architecture benefits from increased capacity, as perplexity generally decreases with more parameters, dropping to 2.18 for the 160M Synthetic model. Furthermore, the inclusion of the "RxT-Alpha Micro (updated)" model, which was trained with improvements to the pipeline, isolates the impact of the training methodology itself. Its notable performance gain (PPL dropping from 2.56 to 2.31) underscores that methodological refinements are as crucial as architectural design.

\subsection{Conversational Coherence Evaluation}

While perplexity measures fluency, the MRL Reward Score provides a more holistic assessment of a model's ability to maintain a coherent, context-aware conversation. Table \ref{tab:mrl_results} presents the benchmark results over an 8+1 step interaction sequence.

\begin{table}[h]
  \caption{MRL Conversational Coherence Benchmark Results (8+1 Interaction Steps).}
  \label{tab:mrl_results}
  \centering
  \begin{tabular}{lcccc}
    \toprule
    \textbf{Model} & \textbf{Parameters} & \textbf{Mean Reward} & \textbf{Max Reward} & \textbf{Min Reward} \\
    \midrule
    LLM Baseline & 22M & 2.4 & 4.0 & 1.4 \\
    \midrule
    RxT-Alpha Nano & 12M & 3.1 & 4.4 & 1.4 \\
    RxT-Alpha Micro (updated) & 26M & 3.4 & 5.2 & 1.8 \\
    RxT-Alpha Mini & 100M & 3.7 & 6.2 & 1.8 \\
    RxT-Alpha Synthetic & 160M & 3.8 & 6.8 & 1.9 \\
    \bottomrule
  \end{tabular}
\end{table}

\begin{figure}
    \centering
    \includegraphics[width=1\linewidth]{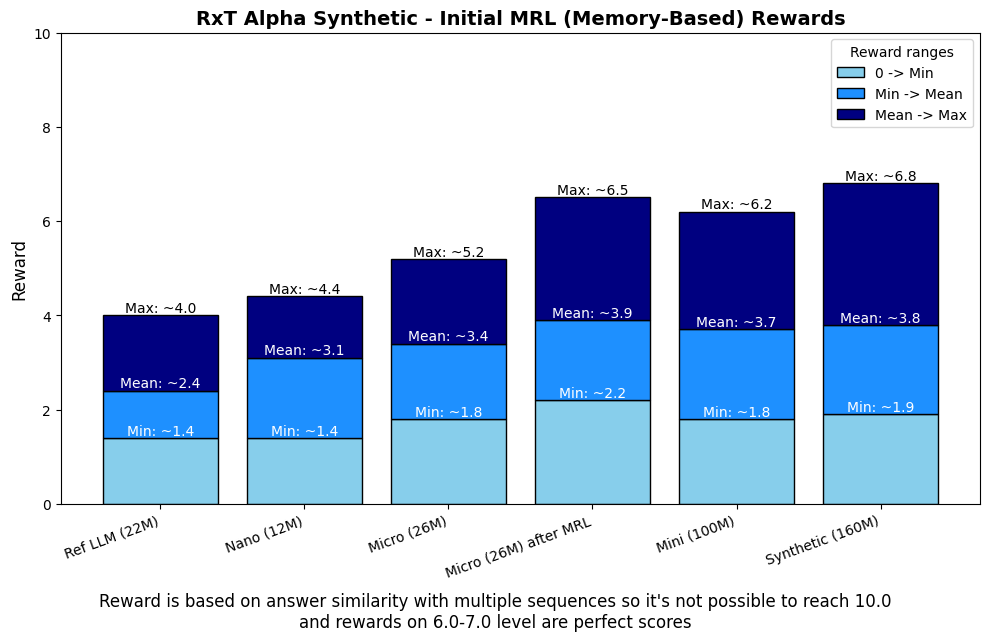}
    \caption{Memory and Dialog Quality Benchmark scores}
    \label{fig:placeholder}
\end{figure}

The MRL reward data reinforces the findings from the perplexity analysis. The mean reward, which serves as the best single indicator of overall conversational quality, shows a clear and consistent trend: all RxT models outperform the baseline, and performance scales with model size, rising from 3.1 for Nano to 3.8 for Synthetic. This provides strong evidence that the memory system is not only functional but also increasingly effective at larger scales.

The minimum and maximum reward scores offer further insight. The higher minimum reward for the RxT models (e.g., 1.8-1.9 for the larger variants vs. 1.4 for the LLM) suggests greater robustness. The dedicated memory system appears to provide a more stable contextual foundation, making the models less prone to catastrophic failures where they completely lose the thread of the conversation. The maximum reward indicates the model's peak performance, showing that larger RxT models are capable of producing significantly higher-quality, more coherent responses.

\subsection{Prompt Phase Latency in Memory-Based Dialogue Benchmark}
\begin{figure}
    \centering
    \includegraphics[width=0.75\linewidth]{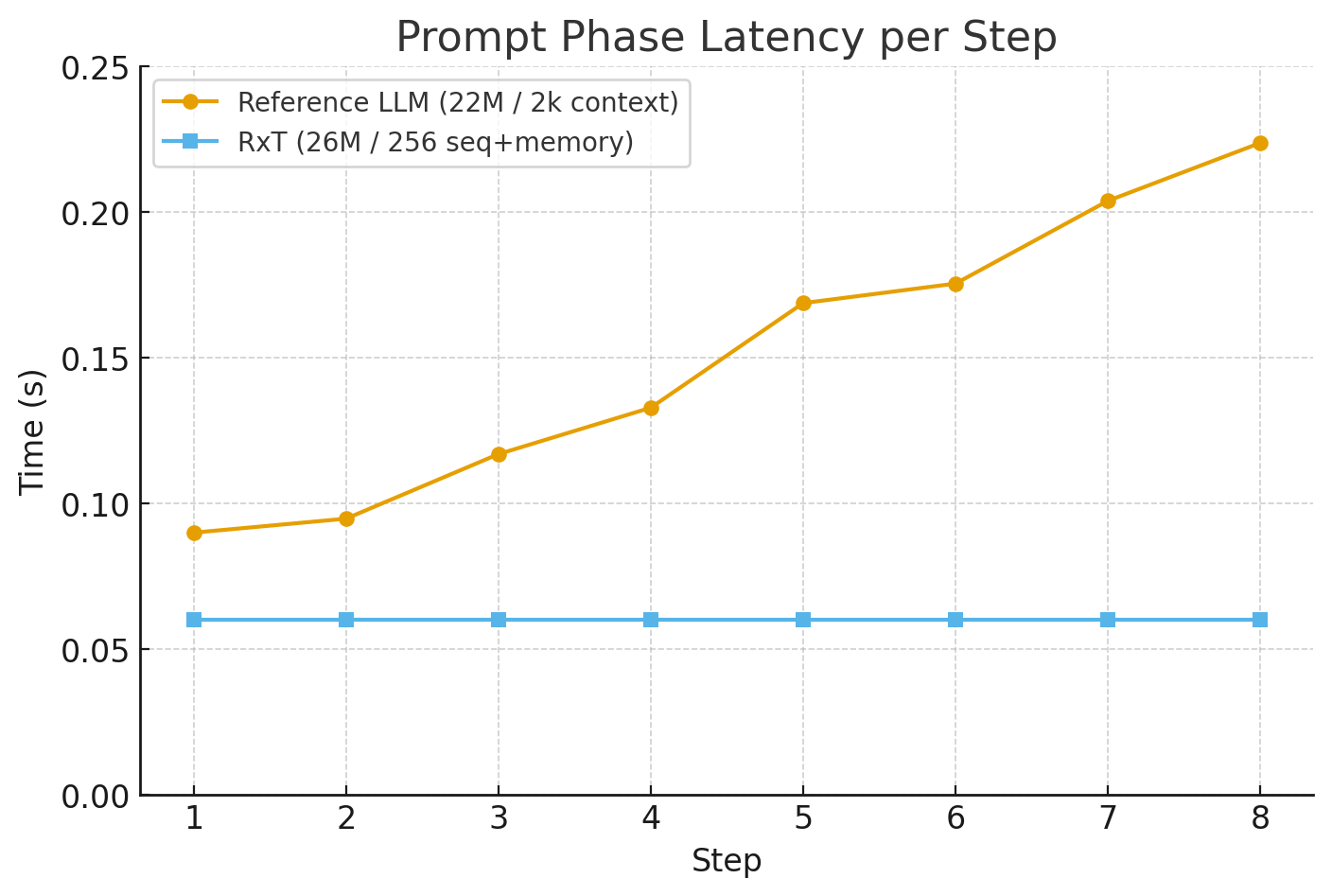}
    \caption{Prompt phase latency in dialog for RxT and reference stateless LLM}
    \label{fig:prompt_latency}
\end{figure}

We benchmarked the prompt phase latency of our RxT model against a stateless reference LLM. 
The benchmark was conducted in the dialogue memory setting with up to 8 conversational steps. 
As shown in Figure~\ref{fig:prompt_latency}, the reference LLM exhibits a steady increase in latency from $0.09$s at step~1 to over $0.22$s at step~8, due to the quadratic dependence on context length inherent to decoder-only architectures. 
In contrast, RxT maintains nearly constant latency across all steps ($\sim$0.06s), independent of dialogue depth, thanks to its fixed-size memory mechanism. 

These results highlight a key efficiency advantage of RxT: unlike decoder-only LLMs, whose prompt phase grows more expensive as the dialogue context expands, RxT achieves predictable and stable inference time, which is critical for interactive applications requiring long-term consistency and low-latency responses.

\subsection{Analysis and Discussion}

The combined experimental results provide compelling evidence for the principle of architectural specialization. The 22M parameter LLM baseline must solve the problem of dialogue context using only its general-purpose self-attention mechanism over a long, undifferentiated sequence of tokens. This is an inefficient, brute-force approach that requires the model to re-discover the dialogue's structure and identify relevant historical information from scratch at every single turn.

In contrast, the 26M RxT-Alpha Micro model, with a comparable parameter count, dramatically outperforms the baseline. This is not a result of more parameters, but of a more intelligent organization of those parameters for the specific task of dialogue. The RxT architecture embodies a "division of labor": the Memory Encoder's role is to summarize the immediate past, the Memory Attention network's role is to integrate that summary into a persistent state, and the Generator-Decoder's role is to generate a response conditioned on this curated context. This design offloads the cognitive burden of long-term context management from the decoder into specialized, purpose-built components.

The benchmarks confirmed the results achieved with MemBART \cite{wu2022stateful}, where memory-augmented transformer significantly outperformed similar stateless alternatives.

The implication is that for complex, structured tasks like maintaining conversational state, simply increasing the size of a generic, monolithic model may be a fundamentally suboptimal path. Designing architectures that reflect the inherent structure of the problem can lead to superior performance with greater parameter and computational efficiency. The linear-time inference cost of RxT is not merely an optimization; it is a symptom of a more appropriate and effective architectural design for stateful interaction. The success of these small-scale models serves as a strong proof-of-concept, motivating future work in scaling the Reactive Transformer to larger models and more complex, real-world datasets.

\section{Conclusion and Future Work}

The Reactive Transformer presents a new architectural paradigm for conversational AI. By embracing an event-driven, stateful approach, it directly addresses the critical bottlenecks of computational complexity and latency that plague current Large Language Models in dialogue applications. Its asynchronous operational cycle and integrated attention-based memory lead to linear scaling of costs and enable genuine real-time interaction.

This work lays the foundation for a new class of Reactive Language Models (RxLMs). Future work will detail the multi-stage training curriculum required to effectively train this architecture, including supervised and reinforcement learning phases. Furthermore, the current Short-Term Memory system is a stepping stone towards more advanced models incorporating a persistent Long-Term Memory (LTM), enabling true live learning and infinite context retention. We believe this direction is essential for moving beyond language modeling and toward developing more capable, aware, and truly interactive AI systems.

Supervised training of Reactive Transformer is just a first stage of bigger curriculum, that's extended by follow-up Reinforcement Learning stages (Memory Reinforcement Learning and Reinforcement Learning from Human Feedback for Reactive Models), that will be deeply described in our future work.

Future work will focus on scaling the Reactive Transformer to larger parameter counts and benchmarking on complex, real-world datasets. A key priority will be a direct comparison against other leading efficient architectures, such as State Space Models like Mamba , to comprehensively map the landscape of next-generation sequence models.

\section*{License and Patent Notice}
Patent pending for the Reactive Transformer architecture (\#P.453260). Commercial usage is regulated by the Reactive AI Models \& Architecture License, that require reactive models to be trained using our RxNN/RxLM framework (https://github.com/RxAI-dev/rxlm) or third party libraries licensed by Reactive AI.

\bibliographystyle{unsrt}

\end{document}